\documentclass{article}

\PassOptionsToPackage{square,numbers}{natbib}
\PassOptionsToPackage{table,xcdraw,dvipsnames}{xcolor}
\usepackage[preprint]{ail_at_hku}

\usepackage[utf8]{inputenc}
\usepackage[T1]{fontenc}
\DeclareUnicodeCharacter{2011}{-}

\usepackage{amsmath,amssymb,amsfonts}
\usepackage{graphicx}
\usepackage{booktabs}
\usepackage{multirow}
\usepackage{enumitem}
\usepackage{pifont}
\usepackage{tabularx}
\usepackage{subcaption}
\usepackage{float}
\usepackage{wrapfig}
\usepackage{array}
\usepackage{url}
\usepackage{hyperref}
\hypersetup{colorlinks=true,citecolor=blue,linkcolor=blue,urlcolor=blue,
  pdftitle={HiMem-WAM: Hierarchical Memory-Gated World Action Models for Robotic Manipulation}}

\newcommand{\method}{HiMem-WAM}

\makeatletter
\renewcommand{\absfooter}[4]{%
  \vskip 8pt
  \begin{minipage}[b]{0.75\linewidth}
    \raggedright
    \small\sffamily
    \if\relax\detokenize{#1}\relax\else\textbf{Code:} #1\par\fi
    \if\relax\detokenize{#2}\relax\else\textbf{Website:} #2\par\fi
    \if\relax\detokenize{#3}\relax\else\textbf{Correspondence:} #3\par\fi
  \end{minipage}%
  \hfill
  \begin{minipage}[b]{0.22\linewidth}
    \raggedleft
    \if\relax\detokenize{#4}\relax\else{\sffamily\textbf{#4}}\fi
  \end{minipage}%
}
\makeatother
\title{HiMem-WAM: Hierarchical Memory-Gated World Action Models for Robotic Manipulation}

\author{%
  \begin{minipage}{\linewidth}
    \centering
    \normalsize\sffamily\bfseries
    Xiaoquan Sun\textsuperscript{2,3,$\dagger$}
    \quad
    Ruijian Zhang\textsuperscript{3}
    \quad
    Chen Cao\textsuperscript{1}
    \quad
    Yihan Sun\textsuperscript{3}
    \quad
    Jiahui Chen\textsuperscript{1,2}
    \quad
    Zetian Xu\textsuperscript{1}
    \quad
    Bo Chen\textsuperscript{1,2}
    \quad
    Haijier Chen\textsuperscript{2,5}
    \quad
    Zhen Yang\textsuperscript{1,2}
    \quad
    Jiarun Zhu\textsuperscript{6}
    \quad
    Yijun Hong\textsuperscript{6}
    \quad
    JingZhe Xu\textsuperscript{3}
    \quad
    Jingrui Pang\textsuperscript{4}
    \quad
    Mingqi Yuan\textsuperscript{1,2,$*$}
    \quad
    Jiayu Chen\textsuperscript{1,2,$*$}
    \\[8pt]
    \small\rmfamily\normalfont
    \makebox[\linewidth][c]{%
      \textsuperscript{1}The University of Hong Kong
      \hspace{0.5em}
      \textsuperscript{2}INFIFORCE
      \hspace{0.5em}
      \textsuperscript{3}Huazhong University of Science and Technology
    }%
    \\[3pt]
    \makebox[\linewidth][c]{%
      \textsuperscript{4}Tsinghua University
      \hspace{0.5em}
      \textsuperscript{5}Wuhan University
      \hspace{0.5em}
      \textsuperscript{6}Southern University of Science and Technology
    }%
    \\[8pt]
    \centering
    \small\rmfamily\normalfont
    \textsuperscript{$\dagger$}Project lead
    \quad
    \textsuperscript{$*$}Corresponding authors
  \end{minipage}
}

\begin{document}
\maketitle

\begin{abstract}{}
                  {}
                  {Mingqi Yuan (my017@hku.hk), Jiayu chen (jiayuc@hku.hk)}
                  {June, 2026}
World Action Models (WAMs) have emerged as a new powerful paradigm for embodied intelligence, learning action-relevant visual dynamics that significantly enhance generalization and robustness. However, existing WAMs still struggle with task-relevant memory in long-horizon robotic manipulation. To address this, we present \textbf{\method}, a \textbf{Hi}erarchical \textbf{Mem}ory-Gated WAM that integrates motion-centric latent actions, high-level skill latents, and boundary-triggered memory updates. Specifically, we develop a hierarchical latent action framework that jointly learns low-level motion and high-level skill latents, providing structured temporal abstraction. Meanwhile, a boundary-aware memory gate writes compact task states at predicted skill transitions, enabling causal inference without test-time generation of future video or optical flow estimation. Evaluated on LIBERO, LIBERO‑PLUS, RMBench and real‑world tasks, HiMem‑WAM shows that hierarchical latents improve robustness under deployment perturbations, and the memory module substantially benefits memory‑dependent long‑horizon manipulation.
\end{abstract}

\begin{figure}[!t]
    \centering
    \includegraphics[width=1.0\textwidth,height=0.99\textheight,keepaspectratio]{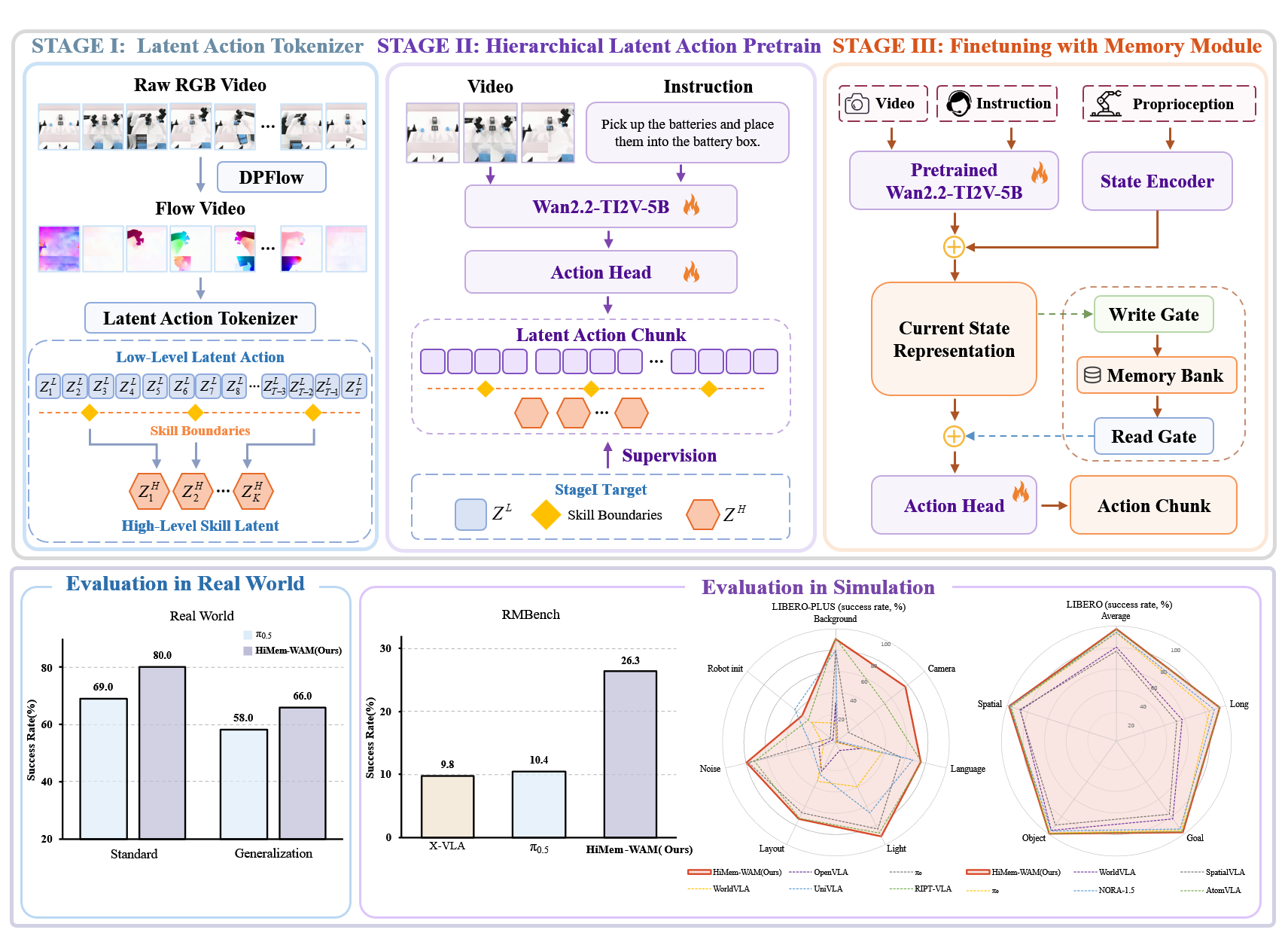}
    \caption{
\textbf{\method\ framework.} \method\ contains three stages: Stage I extracts low-level action tokens and high level skill latents from demonstrations. Stage II learns to predict latent action from video and language inputs. Stage III introduces a gated memory module for history aware action prediction. The bottom panels show real world and simulation evaluations results. 
}
\label{fig:HiMemWAM_Overview}
\end{figure}

\section{Introduction}
Recent Vision-Language-Action (VLA) models have advanced language-conditioned robotic manipulation by transferring semantic priors from large-scale vision-language pre-training to control policies~\citep{rt2,kim2024openvla,octo_2023,gr1,rdt,pi0,pi05,x-vla,llava-vla}. However, these models typically rely on direct end-to-end action prediction, which often lack robustness to deployment shifts ({\em e.g.}, changes in lighting or camera view) and struggle to maintain relevant task history over long periods. World Action Models (WAMs) offer a complementary approach by learning action-relevant visual dynamics through future prediction, video generation, or latent dynamics modeling~\citep{du2023universal,zhou2024robodreamer,feng2025vidar,lv2025f1,zhu2025uwm,motus,worldvla,fastwam}. While these models provide strong dynamics priors and demonstrate superior robustness, they remain inefficient in long-horizon manipulation tasks. 

To enhance the long‑horizon manipulation of WAMs, recent advancements include unified video‑action architectures and verification‑based adaptive execution. For instance, MotuBrain \citep{team2026motubrain} jointly learns video generation and action prediction via a mixture of transformers, enabling autoregressive rollout for extended execution. LingBot‑VA \citep{lingbot-va} interleaves video and action tokens in a causal sequence using KV‑cache for persistent memory, excelling at complex, long-horizon tasks such as preparing breakfast. In parallel, \citep{wang2026trust} proposes a FFDC approach that verifies whether the WAM‑imagined future remains consistent with real observations, dynamically adjusting chunk length, significantly reducing inference cost on the RoboTwin benchmark while improving success rate. However, two important aspects remain underexplored in existing WAMs for long‑horizon manipulation: hierarchical abstraction of low‑level motions into reusable skills, and the ability to explicitly retain memory of task-relevant states across skill boundaries. These capabilities become essential when the robot must recall occluded subtask states or adapt after partial task completion.
In this paper, we present \textbf{\method}, a hierarchical memory-gated WAM for robotic manipulation. \method\ augments the standard WAM pipeline with two key innovations: a hierarchical latent action framework that decomposes long-horizon tasks into reusable skills, and a memory-gated module that retains task-relevant information across skill boundaries. Our main contributions are as follows:
\begin{itemize}[leftmargin=*]
    \item We introduce a two-level latent architecture that organizes low-level motion-centric primitives into high-level skill latents. A planner predicts the current skill from observations, language, proprioception, and memory. An executor then expands the predicted skill into a low-level latent action chunk. A decoder finally maps the chunk to executable robot controls. This structured representation bridges short-horizon execution with long-horizon task decomposition, enabling reasoning at both motion and skill levels.

    \item We develop a memory-gated module that writes compact task states only at predicted skill transitions. A read gate retrieves historical context from an external memory bank to condition the planner, while a write gate uses the predicted boundary score to decide when to store a new memory token. This event-driven design retains occluded subtask states and adapts after partial completion without dense history aggregation. Crucially, as memory updates are triggered by learned boundaries rather than future prediction, \method\ maintains fully causal inference without test-time video generation or optical flow.

    \item  We evaluate \method\ on LIBERO~\citep{libero}, LIBERO-PLUS~\citep{libero-plus}, RMBench~\citep{RMBench}, and real-world tasks. \method\ achieves 97.7\% on LIBERO, 76.0\% on Zero-Shot LIBERO-PLUS, and 26.3\% on RMBench, while outperforming $\pi_{0.5}$\cite{pi05} by an average of 22.5\% on hard real-world tasks. These results demonstrate that \method\ improves robustness under deployment perturbations and delivers consistent gains on long-horizon, memory-dependent tasks.
\end{itemize}

\vspace{-0.8em}

\section{Related Work}
\noindent\textbf{Vision-Language-Action models.}
Vision-Language-Action (VLA) models unify visual observations, language instructions, and robot actions within a single policy interface, enabling language-conditioned manipulation across diverse tasks and embodiments~\citep{rt2,kim2024openvla,octo_2023,gr1,rdt,pi0,pi05,x-vla,llava-vla}. Built on pretrained vision-language representations and large-scale robot demonstrations, these models have shown strong potential for generalist robot control. Action-chunking and diffusion-based policies further improve execution by predicting temporally coherent action sequences rather than single-step controls~\citep{act,dp}. 

\noindent\textbf{World Action Models.}
World models learn action-relevant representations by predicting future observations, latent dynamics, or imagined rollouts~\citep{du2023universal,zhou2024robodreamer,feng2025vidar,lv2025f1}. Recent World Action Models connect visual prediction and action generation within a unified framework~\citep{zhu2025uwm,worldvla,lingbot-va,motus,fastwam,cogact,cronusvla}. Motus~\citep{motus} shows that latent actions can bridge action-free videos and robot demonstrations, while Fast-WAM~\citep{fastwam} studies whether future imagination is necessary at test time.

\noindent\textbf{Manipulation policies with memory.}
Long horizon manipulation often requires recalling information that is no longer visible, such as object states, completed subtasks, or task progress. RMBench~\citep{RMBench} highlights this challenge and shows the limitation of fixed length observation histories for retaining task relevant information. Recent methods, including MemoryVLA~\citep{memoryvla}, MemER~\citep{memer}, SAM2Act~\citep{sam2act}, and CronusVLA~\citep{cronusvla}, introduce memory or multi frame reasoning for long horizon decision making. Unlike dense history aggregation or fixed memory windows, \method\ writes compact task states only at learned skill boundaries and uses them for memory dependent action prediction.

\begin{figure*}[t]
    \centering
    \includegraphics[width=1.0\textwidth]{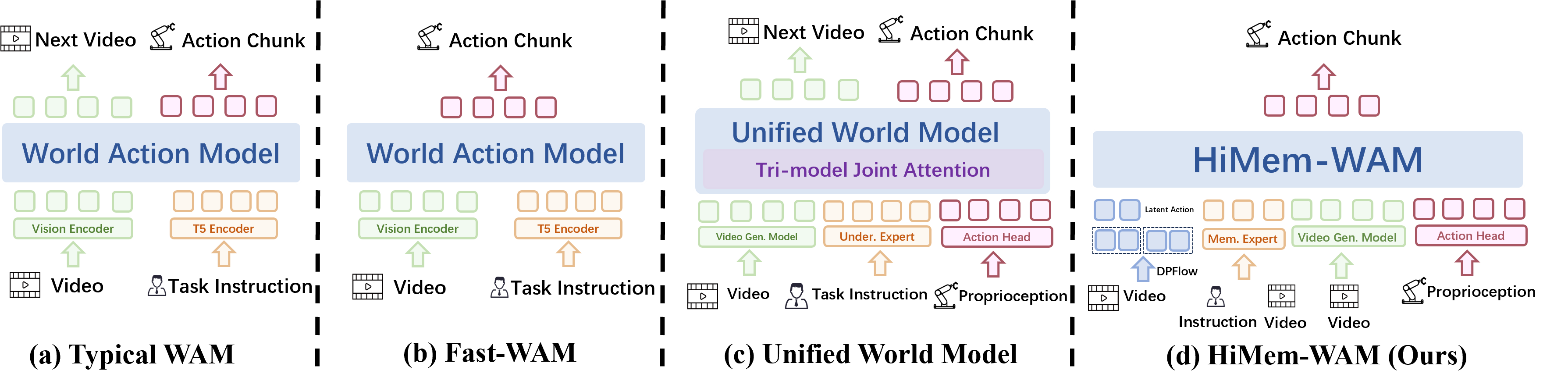}
    \caption{
    \textbf{From WAM to HiMem-WAM.}
    \method\ extends unified world action modeling with a memory expert, enabling action prediction conditioned on both current observations and task history.
    }
    \label{fig:architecture_comparison}
\end{figure*}

\section{Method}
\label{sec:method}
\noindent\textbf{Overview.}
We study long-horizon manipulation with multi-view observations. At timestep $t$, the policy receives RGB observations $o_t=\{I_t^{(v)}\}_{v=1}^{V}$, proprioception $p_t$, a task instruction $\ell$, and an external memory state $\mathcal{M}_t$. It predicts an action chunk $\mathbf{a}_{t:t+K-1}=(a_t,\ldots,a_{t+K-1})$. The central design of \method\ is to factor this policy through a high-level skill latent $z_t^h$ and a low-level latent-action chunk $\mathbf{Z}_{t:t+K-1}^l=(z_t^l,\ldots,z_{t+K-1}^l)$:
\begin{equation}
\begin{aligned}
\pi_\theta(\mathbf{a}_{t:t+K-1}\mid o_t,p_t,\ell,\mathcal{M}_t)
&=\int p_\theta(\mathbf{a}_{t:t+K-1}\mid \mathbf{Z}_{t:t+K-1}^l,o_t,p_t) \\
&\quad\cdot p_\theta(\mathbf{Z}_{t:t+K-1}^l\mid z_t^h,o_t,p_t,\mathcal{M}_t) \\
&\quad\cdot p_\theta(z_t^h\mid o_t,p_t,\ell,\mathcal{M}_t)\,d\mathbf{Z}^l\,dz^h .
\end{aligned}
\label{eq:hierarchical_policy}
\end{equation}
The factorization separates three roles: selecting the current skill, unfolding the skill into short-horizon motion, and grounding that motion into embodiment-specific controls. Future visual dynamics are used only as training supervision, so inference remains causal and does not require video generation or optical-flow estimation.

\noindent\textbf{Low-level latent actions.}
Low-level latent actions provide a compact motion space that can be learned from both action-labeled robot trajectories and action-free videos. For each transition, we compute multi-view optical flow $\mathbf{\Phi}_t=\{\Phi_t^{(v)}\}_{v=1}^{V}$ with DPFlow~\cite{DPFlow}. A variational tokenizer encodes the short-horizon context $c_t=(o_t,o_{t+1},p_t,\ell,\mathbf{\Phi}_t)$ into
\begin{equation}
q_\phi(z_t^l\mid c_t)=\mathcal{N}(\mu_t,\mathrm{diag}(\sigma_t^2)),
\qquad
z_t^l=\mu_t+\sigma_t\odot\epsilon,
\quad \epsilon\sim\mathcal{N}(0,I).
\label{eq:low_latent_posterior}
\end{equation}
The tokenizer is trained to reconstruct visual motion and, when action labels are available, weakly align the latent space with real controls:
\begin{equation}
\mathcal{L}_{l}
=
\|\hat{\mathbf{\Phi}}_t-\mathbf{\Phi}_t\|_1
+
\lambda_a\mathbb{I}^{\mathrm{act}}_t\|\hat a_t-a_t\|_2^2
+
\beta D_{\mathrm{KL}}\!\left(q_\phi(z_t^l\mid c_t)\,\|\,\mathcal{N}(0,I)\right).
\label{eq:low_latent_loss}
\end{equation}
Here $\mathbb{I}^{\mathrm{act}}_t$ indicates whether the transition has an action label, $D_{\mathrm{KL}}$ denotes Kullback-Leibler (KL) divergence, $\hat{\mathbf{\Phi}}_t$ is the reconstructed flow, and $\hat a_t$ is an auxiliary action prediction. After training, the tokenizer is frozen and applied offline to all trajectories and videos to produce the low-level sequence $Z^l=(z_1^l,\ldots,z_{T-1}^l)$.

\noindent\textbf{High-level skill latents.}
Low-level latents capture local motion, but long-horizon tasks require a coarser notion of progress. We therefore learn high-level latent actions, or skill latents, by dynamically chunking $Z^l$. Let $Z^{(0)}=Z^l$. At hierarchy stage $s$, an encoder maps each token to $h_i^{(s)}$, a boundary predictor marks the starts of new segments with $b_i^{(s)}$, and a segment pooling module summarizes each variable-length segment into the next-stage token:
\begin{equation}
Z^{(s+1)}=\mathrm{Chunk}_s(E_s(Z^{(s)});b^{(s)}),
\qquad
Z^h=Z^{(H)}.
\label{eq:skill_chunk_main}
\end{equation}
The learned boundaries define where one skill ends and the next begins. For policy supervision, the final skill sequence $Z^h=(z_1^h,\ldots,z_S^h)$ is unfolded back to the original control timeline, yielding a per-timestep skill target $\bar z_t^h$ and a boundary label $\bar b_t$. Full boundary scoring, pooling, and unfolding details are given in Appendix~\ref{app:skill_unfolding}.

\noindent\textbf{Memory-gated Module.}
Long-horizon manipulation often depends on observations that are no longer visible. \method\ therefore uses a gated memory adapter that stores compact skill-level events. We first encode the current state as $x_t=E_\theta(o_t,p_t,\ell)$ and retrieve a memory context
\begin{equation}
c_t^m=\mathrm{Attn}(W_qx_t,W_k\mathcal{M}_t,W_v\mathcal{M}_t),
\qquad
\tilde{x}_t=x_t+\alpha_t^r W_m c_t^m,
\qquad
\alpha_t^r=\sigma(G_r(x_t,c_t^m)).
\label{eq:memory_read_main}
\end{equation}
The scalar $\alpha_t^r$ is the read gate, which controls how much retrieved memory is injected into the state. A Qwen3-VL-4B-Instruct planner then predicts the current skill and a boundary score,
\(
(\hat z_t^h,\hat b_t)=\pi_\theta^{\mathrm{plan}}(\tilde{x}_t,c_t^m),
\)
and the executor predicts the low-level latent-action chunk,
\(
\hat{\mathbf{Z}}_{t:t+K-1}^l=\pi_\theta^{\mathrm{exec}}(\tilde{x}_t,\hat z_t^h).
\)
An action decoder maps this chunk to executable controls:
\(
\hat{\mathbf{a}}_{t:t+K-1}=D_{\mathrm{act}}(\hat{\mathbf{Z}}_{t:t+K-1}^l,\tilde{x}_t).
\)
Memory writing is controlled by a write gate,
\begin{equation}
\alpha_t^w=\sigma(G_w(\tilde{x}_t,\hat z_t^h,\hat b_t)),
\qquad
\mathcal{M}_{t+1}=\begin{cases}
U_\psi(\mathcal{M}_t,\gamma_t), & \alpha_t^w>\eta,\\
\mathcal{M}_t, & \mathrm{otherwise}.
\end{cases}
\label{eq:memory_write_main}
\end{equation}
where $\gamma_t$ is a candidate memory token formed from the current state, predicted skill, and pooled low-level latent chunk. This boundary-aware gate prevents dense memory growth and aligns memory updates with meaningful task transitions.

\noindent\textbf{Training Pipeline.}
Training proceeds in three stages.

\noindent\emph{Stage I: Latent Action Tokenizer.}
We first learn the low-level tokenizer with Eq.~\eqref{eq:low_latent_loss}.

\noindent\emph{Stage II: Hierarchical Latent Action Pretrain.}
We freeze the tokenizer, discover skill latents, and pretrain the planner and executor without external memory. 
The Stage-II objective combines Mean-Squared Error (MSE) losses for the predicted skill and latent action chunk with binary cross-entropy (BCE) for boundary prediction:
\begin{equation}
\mathcal{L}_{\mathrm{latent}}
=
\lambda_h\mathrm{MSE}(\hat z_t^h,\bar z_t^h)
+
\lambda_l\mathrm{MSE}(\hat{\mathbf{Z}}_{t:t+K-1}^l,\mathbf{Z}_{t:t+K-1}^l)
+
\lambda_b\mathrm{BCE}(\hat b_t,\bar b_t).
\label{eq:latent_pretrain}
\end{equation}

\noindent\emph{Stage III: Finetuning with Memory-gated Module.}
We activate memory and fine-tune the full policy on action-labeled demonstrations:
\begin{equation}
\mathcal{L}_{\mathrm{ft}}
=
\mathcal{L}_{\mathrm{act}}
+
\alpha_h\mathcal{L}_{\mathrm{plan}}
+
\alpha_l\mathcal{L}_{\mathrm{exec}}
+
\alpha_b\mathcal{L}_{\mathrm{bd}}
+
\alpha_m\mathcal{L}_{\mathrm{gate}}.
\label{eq:finetune_main}
\end{equation}
where $\mathcal{L}_{\mathrm{act}}$ is either an action MSE for deterministic policies or a negative log-likelihood for stochastic policies. 
The gate loss is
\begin{equation}
\mathcal{L}_{\mathrm{gate}}
=
\mathrm{BCE}(\alpha_t^w,\bar b_t)
+
\lambda_r\|\alpha_t^r\|_1
+
\lambda_w\|\alpha_t^w\|_1.
\label{eq:gate_loss_main}
\end{equation}
BCE supervises the write gate with discovered skill-boundary labels, while the $\ell_1$ terms encourage sparse memory reading and writing.

\noindent\textbf{Inference.}
At test time, \method\ uses only the current RGB observations, proprioception, instruction, and memory bank. It reads memory, predicts a high-level latent action, expands it into a low-level latent-action chunk, decodes the chunk into executable actions, and writes a new memory token only when $\alpha_t^w>\eta$. The procedure is fully causal and preserves the standard action-chunking interface used by robot policies.

\section{Experiments}
We design our experiments to evaluate whether \method\ improves robotic manipulation performance. In particular, we focus on the following questions:
\begin{itemize}[leftmargin=*]
    \item \textbf{Q1:} How does \method\ compare with existing VLAs and WAMs on standard manipulation benchmarks?
    \item \textbf{Q2:} Do latent actions provide effective motion priors for robotic manipulation?
    \item \textbf{Q3:} Does the gated-memory module enhance long-horizon memory in robotic manipulation?
    \item \textbf{Q4:} Can \method\ be effectively deployed on real-world robotic tasks?
\end{itemize}

\subsection{Simulation Experimental Setup}
\noindent\textbf{Benchmark Selection.}
We evaluate \method\ on three benchmarks: LIBERO~\citep{libero}, LIBERO-PLUS~\citep{libero-plus}, and RMBench~\citep{RMBench}. LIBERO~\cite{libero} evaluates language-conditioned manipulation across four task suites, including spatial, object, goal, and long-horizon tasks. LIBERO-PLUS~\citep{libero-plus} evaluates policy robustness under deployment-time perturbations across vision, language, initialization, and scene layout. RMBench~\cite{RMBench} evaluates memory-dependent manipulation tasks that require the agent to retain and reuse task-relevant information across long horizons.

\noindent\textbf{Evaluation Metrics.}
We use \textbf{S}uccess \textbf{R}ate~(\textbf{SR}) as the evaluation metric across all benchmarks. On LIBERO~\cite{libero} and LIBERO-PLUS~\cite{libero-plus}, we evaluate 50 rollouts per task. On RMBench~\cite{RMBench}, we evaluate 100 rollouts per task.

\noindent\textbf{Algorithmic Baselines.}
We evaluate \method\ against a collection of representative and competitive baselines, including DP~\cite{dp}, ACT~\cite{act}, $\pi_{0.5}$~\cite{pi05}, OpenVLA~\cite{kim2024openvla}, X-VLA~\cite{x-vla}, MEM-0~\cite{RMBench}, AtomVLA~\cite{atomvla}, WorldVLA~\cite{worldvla}, LingBot-VA~\cite{lingbot-va}, Fast-WAM~\cite{fastwam} and other baselines.

\subsection{Results Analysis}

\noindent\textbf{Key Finding 1: Latent actions improve robustness by capturing transferable motion patterns.}
As shown in Tables~\ref{tab:libero_benchmark} and~\ref{tab:libero_plus_benchmark}, \emph{Stage II} latent action pretraining achieves gains of \textbf{+1.1\%} on the standard LIBERO benchmark and \textbf{+3.8\%} on the LIBERO-PLUS benchmark (Zero-Shot), where models are trained only on the standard LIBERO dataset. The larger gain under deployment perturbations, including camera and viewpoint variations and observation noise, suggests that latent actions capture task completion motion rather than overfitting to clean visual trajectories. This provides a stable motion prior for action prediction and improves robustness under diverse deployment perturbations.
\begin{table}[H]
\vspace{-4mm}
\centering
\captionsetup{skip=1pt}
\caption{Comparisons with other baselines on RMBench~\cite{RMBench} benchmark.}
\label{tab:rmbench_results}
\scriptsize
\setlength{\tabcolsep}{3pt}
\renewcommand{\arraystretch}{0.76}
\setlength{\aboverulesep}{0.2ex}
\setlength{\belowrulesep}{0.2ex}

\resizebox{\linewidth}{!}{%
\begin{tabular}{@{}lc|ccccc@{}}
\toprule
Tasks & TMC & DP & ACT & \( \pi_{0.5} \) & X-VLA & \textbf{\method~(Ours)}\\
\midrule
\textit{Observe and Pick Up}  & $M(1)$ & 1\%  & 1\%  & 9\%  & 9\%  & \textbf{28\%}\\
\textit{Rearrange Blocks}    & $M(1)$ & 0\%  & 29\% & 13\% & 13\% & \textbf{33\%}\\
\textit{Put Back Block}      & $M(1)$ & 0\%  & 0\%  & 11\% & 18\% & \textbf{32\%}\\
\textit{Swap Blocks}         & $M(1)$ & 11\% & 2\%  & 24\% & 16\% & \textbf{38\%}\\
\textit{Swap T}              & $M(1)$ & 2\%  & 2\%  & 15\% & 3\%  & \textbf{27\%}\\
\textbf{Average}             & $M(1)$ & 2.8\% & 6.8\% & 14.4\% & 11.8\% & \textbf{31.6\%}\\
\midrule
\textit{Battery Try}         & $M(n)$ & 10\% & 19\% & 16\% & 26\% & \textbf{28\%}\\
\textit{Blocks Ranking Try}  & $M(n)$ & 10\% & 0\%  & 6\%  & 1\%  & \textbf{24\%}\\
\textit{Cover Blocks}        & $M(n)$ & 0\%  & 0\%  & 2\%  & 0\%  & \textbf{19\%}\\
\textit{Press Button}        & $M(n)$ & 0\%  & 0\%  & 1\%  & 2\%  & \textbf{8\%}\\
\textbf{Average}             & $M(n)$ & 5.0\% & 4.8\% & 6.3\% & 7.3\% & \textbf{19.8\%}\\
\midrule
\textbf{Total Average}       & /      & 3.8\% & 5.9\% & 10.8\% & 9.8\% & \textbf{26.3\%}\\
\bottomrule
\end{tabular}%
}
\vspace{0.2mm}
\noindent\parbox{\linewidth}{\scriptsize
\textit{Note:} We report \textbf{SR} (\%). 
RMBench includes nine manipulation tasks across the $M(1)$ and $M(n)$ levels of Task Memory Complexity~(TMC). 
\textbf{Bold} denotes the best performance among all methods.
}
\vspace{-5mm}
\end{table}
\noindent\textbf{Key Finding 2: Gated memory improves task state tracking, while stronger memory planning is still needed.}
As shown in Table~\ref{tab:rmbench_results}, \method\ achieves a total average SR of \textbf{26.3\%} on RMBench and reaches \textbf{31.6\%} on \(M(1)\) tasks. This shows that the memory module is effective: the read gate retrieves task history to form a memory-adapted state, while the write gate stores compact key states only when meaningful task transitions are detected. As a result, the Qwen3-VL-4B planner can condition action prediction on both the current observation and previously observed task states, which improves target tracking, phase recognition, and memory-dependent manipulation. However, performance drops to \textbf{19.8\%} on \(M(n)\) tasks and remains below Mem-0~\cite{RMBench}, indicating that repeated trials and multi-step state updates still require stronger memory reasoning. This gap is reasonable because Mem-0 uses a larger 8B planner and a more specialized memory planning design, whereas \method\ integrates gated memory into a general world action policy together with latent action priors.

\begin{figure*}[t]
    \centering
    \includegraphics[width=1.0\textwidth]{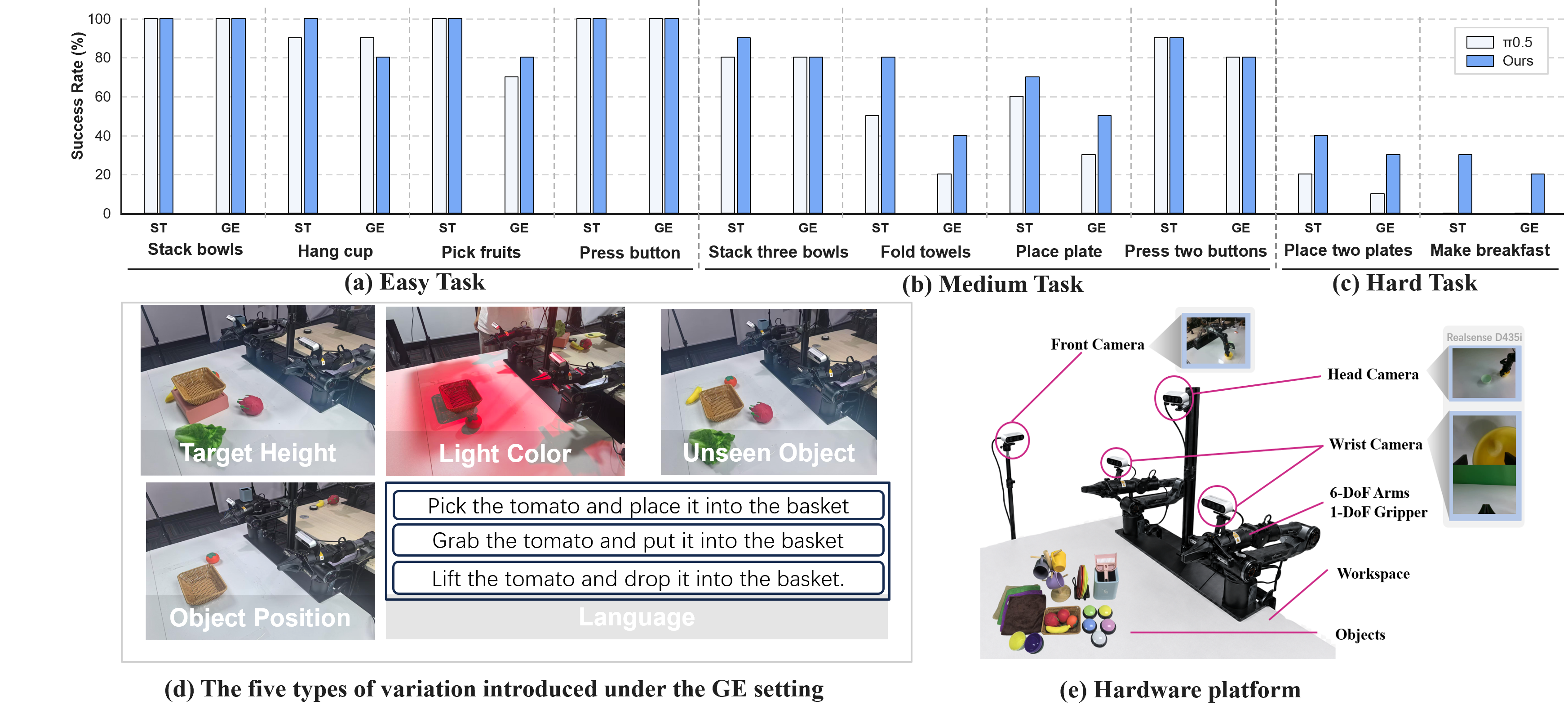}
    \caption{
    \textbf{Real-world evaluation on 10 tasks.}
    We evaluate \method\ on 10 real-world tasks under both the \textbf{ST} and \textbf{GE} settings.
    (a)--(c) report \textbf{SR} across three task categories. (d) illustrates the evaluation variations in the \textbf{GE} setting. (e) illustrates the hardware platform.}
    \label{fig:real_world_results}
\end{figure*}

\subsection{Real-World Experimental Setup}
\noindent\textbf{Hardware Platform.}
We conduct real-world experiments on a dual-arm platform built with two AgileX Piper 6-DoF arms. The system is equipped with four Intel RealSense D435i RGB cameras, including two wrist cameras, one head camera, and one front camera, as shown in Figure~\ref{fig:real_world_results} (e).

\noindent\textbf{Task Settings.} We evaluate real-world deployment on a suite of manipulation tasks organized into three difficulty categories: \textbf{Easy}, \textbf{Medium}, and \textbf{Hard}. Easy tasks focus on short-horizon single-arm manipulation, Medium tasks require basic bimanual coordination, and Hard tasks involve long-horizon bimanual manipulation with more complex object interactions.

\noindent\textbf{Evaluation Settings.} Each task is evaluated under two settings: Standard \textbf{(ST)}, where the scene configuration is clean and consistent with the training data, and Generalization \textbf{(GE)}, where we introduce a range of deployment-time perturbations to evaluate robustness in more challenging environments. We report \textbf{S}uccess \textbf{R}ate~(\textbf{SR}) separately for each difficulty category and evaluation setting. Detailed task definitions and the full specification of the \textbf{GE} setting are provided in Appendix~\ref{app:real_world_tasks}.

\paragraph{Training Details.}
Each task consists of 400 demonstrations. The model undergoes Supervised Fine-Tuning (SFT) on the real-world dataset for 5 epochs. During evaluation, we conduct 20 trials for each task. Specifically, under the GE setting, the environmental conditions of these 20 trials are randomly distributed across the four perturbation types to rigorously assess robustness.

\begin{table*}[t]
\centering
\vspace{-1mm}

\begingroup
\fontsize{6.5pt}{7.1pt}\selectfont
\setlength{\tabcolsep}{1.2pt}
\renewcommand{\arraystretch}{1.03}

\newcommand{\tabgroup}[2]{%
\rowcolor{gray!15}
\multicolumn{#1}{@{}l}{\strut\textbf{#2}} \\
}

\begin{minipage}[t]{0.37\textwidth}
\vspace{0pt}
\centering
\captionsetup{
  type=table,
  font=scriptsize,
  labelfont=normalfont,
  textfont=normalfont,
  justification=raggedright,
  singlelinecheck=false
}
\captionof{table}{Comparisons with other baselines on  LIBERO benchmark.}
\label{tab:libero_benchmark}
\vspace{0.8mm}

\begin{tabular*}{\linewidth}{@{\extracolsep{\fill}}lccccc@{}}
\toprule
\textbf{Method} & \textbf{Spatial} & \textbf{Object} & \textbf{Goal} & \textbf{Long} & \textbf{Avg.} \\
\midrule
\tabgroup{6}{Vision-Language-Action Models}
OpenVLA       & 84.7 & 88.4 & 79.2 & 53.7 & 76.5 \\
SpatialVLA    & 88.2 & 89.9 & 78.6 & 55.5 & 78.1 \\
$\pi_0$       & 96.8 & 98.8 & 95.8 & 85.2 & 94.2 \\
NORA-1.5      & 97.3 & 96.4 & 94.5 & 89.6 & 94.5 \\
AtomVLA       & 96.4 & 99.6 & 97.6 & 94.4 & 97.0 \\
\midrule
\tabgroup{6}{World Action Models}
WorldVLA      & 87.6 & 96.2 & 83.4 & 60.0 & 81.8 \\
Fast-WAM      & \textbf{98.2} & \textbf{100.0} & 97.0 & \textbf{95.2} & 97.6 \\
\midrule
\tabgroup{6}{Ours}
w/o Stage II  & 96.0 & 99.6 & 97.1 & 93.8 & 96.6 \\
\method       & \textbf{98.2} & 99.8 & \textbf{98.4} & 94.5 & \textbf{97.7} \\
\bottomrule
\end{tabular*}
\end{minipage}%
\hfill
\begin{minipage}[t]{0.61\textwidth}
\vspace{0pt}
\centering
\captionsetup{
  type=table,
  font=scriptsize,
  labelfont=normalfont,
  textfont=normalfont,
  justification=raggedright,
  singlelinecheck=false
}
\captionof{table}{Comparisons with other baselines on LIBERO-PLUS benchmark. (Zero-Shot, train on the standard LIBERO dataset only)}
\label{tab:libero_plus_benchmark}
\vspace{0.8mm}

\begin{tabular*}{\linewidth}{@{\extracolsep{\fill}}lcccccccc@{}}
\toprule
\textbf{Method} & \textbf{Cam.} & \textbf{Init} & \textbf{Lang.} & \textbf{Light} & \textbf{BG.} & \textbf{Noise} & \textbf{Layout} & \textbf{Avg.} \\
\midrule
\tabgroup{9}{Vision-Language-Action Models}
OpenVLA       & 0.8  & 3.5  & 23.0 & 8.1  & 34.8 & 15.2 & 28.5 & 16.3 \\
OpenVLA-OFT   & 56.4 & 31.9 & \textbf{79.5} & 88.7 & \textbf{93.3} & 75.8 & 74.2 & 71.4 \\
$\pi_0$       & 13.8 & 6.0  & 58.8 & 85.0 & 81.4 & 79.0 & 68.9 & 56.1 \\
UniVLA        & 1.8  & 46.2 & 69.6 & 69.0 & 81.0 & 21.2 & 31.9 & 45.8 \\
RIPT-VLA      & 55.2 & 31.2 & 77.6 & 88.4 & 91.6 & 73.5 & 74.2 & 70.2 \\
\midrule
\tabgroup{9}{World Action Models}
WorldVLA      & 0.1  & 27.9 & 41.6 & 43.7 & 17.1 & 10.9 & 38.0 & 25.6 \\
HoloBrain-0   & 65.5 & \textbf{58.2} & 78.7 & 88.1 & 90.3 & 66.9 & \textbf{79.5} & 75.3 \\
\midrule
\tabgroup{9}{Ours}
w/o Stage II  & 77.2 & 37.9 & 71.7 & 91.1 & 83.6 & 73.2 & 70.7 & 72.2 \\
\method       & \textbf{78.2} & 38.1 & 76.6 & \textbf{92.2} & 91.0 & \textbf{80.7} & 74.9 & \textbf{76.0} \\
\bottomrule
\end{tabular*}
\end{minipage}

\vspace{0.6mm}
\noindent\parbox{\textwidth}{\scriptsize
\textit{Note:} We report \textbf{SR} (\%). 
\textbf{Bold} denotes the best performance among all methods. 
\textbf{Cam.} denotes camera perturbation; 
\textbf{Init.} denotes initial-state perturbation; 
\textbf{Lang.} denotes language perturbation; 
\textbf{Light} denotes lighting perturbation; 
\textbf{BG.} denotes background perturbation; 
\textbf{Noise} denotes observation noise; 
\textbf{Layout} denotes layout perturbation; 
\textbf{Avg.} denotes average performance.
}

\endgroup

\vspace{-1mm}
\end{table*}

\begin{figure*}[t]
    \centering
    \includegraphics[width=1.0\textwidth]{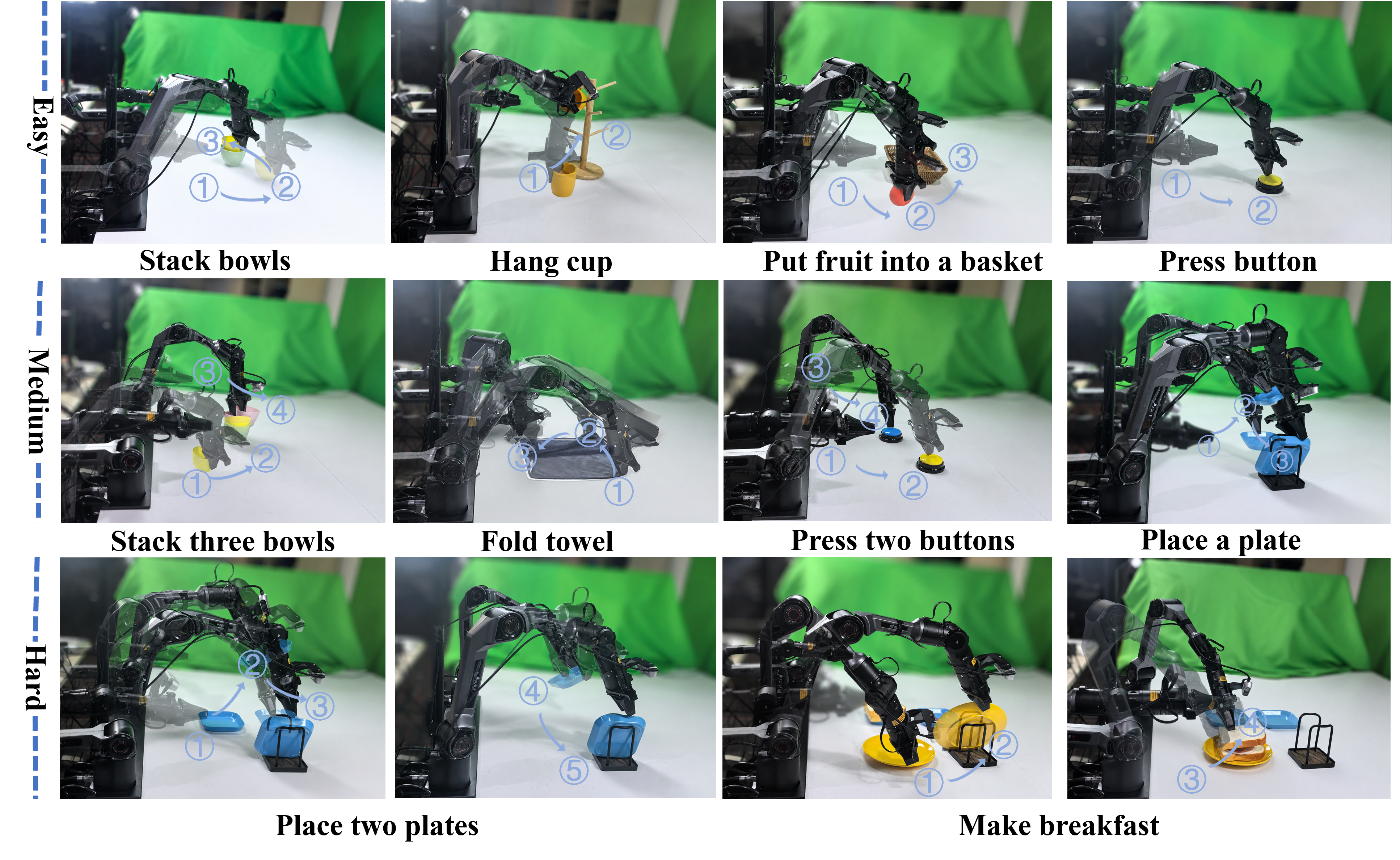}
    \caption{
Visualization of the 10 real-world tasks. 
The first row shows easy tasks:~\emph{Stack bowls, Hang cup, Put fruit into a basket, Press button}, the second row shows medium tasks:~\emph{Stack three bowls, Fold towel, Place plate, Press two buttons}, and the third row shows hard tasks: ~\emph{place two plates, make breakfast.}
}
    \label{fig:visual}
\end{figure*}

\subsection{Results Analysis}

\noindent\textbf{Key Finding 3: \method\ improves generalization in real-world robotic tasks.}
\begin{wraptable}{r}{0.40\linewidth}
\vspace{-2mm}
\centering
\captionsetup{
  font=scriptsize,
  labelfont=normalfont,
  textfont=normalfont,
  justification=raggedright,
  singlelinecheck=false
}
\caption{Real world evaluation under different action representations.}
\label{tab:real_world_category_results}
\vspace{-1mm}

\begingroup
\fontsize{6.3pt}{6.9pt}\selectfont
\setlength{\tabcolsep}{1.6pt}
\renewcommand{\arraystretch}{1.00}

\begin{tabular*}{\linewidth}{@{\extracolsep{\fill}}ccccc@{}}
\toprule
\textbf{Category}
& \multicolumn{2}{c}{\textbf{Joint Pos.}}
& \multicolumn{2}{c}{\textbf{EE Pose}} \\
\cmidrule(lr){2-3} \cmidrule(lr){4-5}
& \textbf{w/o Stage II} & \textbf{w/ Stage II} 
& \textbf{w/o Stage II} & \textbf{w/ Stage II} \\
\midrule
Easy   & 100.0 & 100.0 & 90.0 & 100.0 \\
Medium & 80.0  & 82.5  & 67.5 & 75.0 \\
Hard   & 15.0  & 35.0  & 10.0 & 30.0 \\
\bottomrule
\end{tabular*}

\vspace{0.5mm}
\noindent\parbox{\linewidth}{%
\fontsize{5.5pt}{6.1pt}\selectfont
\textit{Note:} We report \textbf{SR} (\%). All experiments in this table are conducted under the \textbf{ST} setting. \textbf{Joint Pos.} denotes joint-position action representation, \textbf{EE Pose} denotes end-effector pose action representation.
}

\endgroup
\vspace{-3mm}
\end{wraptable}
As shown in Figure~\ref{fig:real_world_results} and Table~\ref{tab:real_world_category_results}, \method\ performs comparably to \(\pi_{0.5}\)~\cite{pi05} on Easy tasks, where both policies already achieve high success rates, confirming that \(\pi_{0.5}\) remains a strong baseline for basic manipulation.
The advantage of \method\ becomes clearer on Medium tasks, with gains of \textbf{+12.5\%} under ST and \textbf{+10.0\%} under GE, suggesting that latent action pretraining provides predictive motion priors that help stabilize real-world action prediction under object, lighting, and instruction variations.
The advantage becomes most pronounced on Hard tasks, with gains of \textbf{+25.0\%} under ST and \textbf{+20.0\%} under GE.
These tasks require long-horizon execution and persistent task-state maintenance, while SFT adaptation often struggles with history forgetting and fails to track the correct task progress.
In contrast, the gated memory module helps \method\ retain historical information and task states, enabling the policy to continue long-horizon task execution.

\noindent\textbf{Key Finding 4: Stage II Latent Action Pretrain is effective across action representations.} 
 As shown in Table~\ref{tab:real_world_category_results}, Stage II hierarchical latent action pretraining improves real-world \textbf{SR} under both Joint Pos. and EE Pose in the ST setting, with larger gains as task difficulty increases. On hard tasks, SR increases from 15.0\% to 35.0\% for Joint Pos. and from 10.0\% to 30.0\% for EE Pose, suggesting that hierarchical latent actions provide useful motion priors and skill level temporal structure for long horizon manipulation. The larger gain for EE Pose indicates that task space commands, which contain less robot-specific execution detail, benefit more from learned motion priors, while the stronger final Joint Pos. result suggests that joint space actions still preserve low-level action information.

\section{Conclusion}
We presented \method, a hierarchical memory-gated WAM for long-horizon manipulation. \method\ combines latent action pretraining, skill latent, and memory-gated module to connect motion execution with task state retention. It learns motion priors from multi-view visual dynamics and uses discovered skill boundaries to control sparse memory writing, enabling causal action prediction from current observations and task history. Experiments in simulation and on real robots show that latent actions improve robustness under deployment perturbations, memory-gated modules support memory-dependent manipulation. These results validate the benefit of coupling latent motion priors with task memory for long-horizon control.

\section{Limitations}
Although \method\ achieves strong results in simulation and real-world manipulation, several limitations remain. First, the training pipeline still requires considerable computation, as it includes optical flow extraction, latent action learning, skill discovery, and memory policy training. Second, the multi-stage design introduces additional engineering complexity, and its performance depends on the quality of learned latent actions, skill boundaries, and memory updates. Finally, our real-world evaluation is currently conducted on a specific dual-arm platform with 10 manipulation tasks. While these experiments cover different difficulty levels and perturbations, broader validation with more tasks, larger data scale, and diverse robot embodiments is needed to further examine the scalability of the framework.

\clearpage

\bibliography{paper_refs}

@article{libero,
  title={LIBERO: Benchmarking Knowledge Transfer for Lifelong Robot Learning},
  author={Liu, Bo and Zhu, Yifeng and Gao, Chongkai and Feng, Yihao and Liu, Qiang and Zhu, Yuke and Stone, Peter},
  journal={arXiv preprint arXiv:2306.03310},
  year={2023}
}

@article{RMBench,
  title={RMBench: Memory-Dependent Robotic Manipulation Benchmark with Insights into Policy Design},
  author={Chen, Tianxing and Wang, Yuran and Li, Mingleyang and Qin, Yan and Shi, Hao and Li, Zixuan and Hu, Yifan and Zhang, Yingsheng and Wang, Kaixuan and Chen, Yue and others},
  journal={arXiv preprint arXiv:2603.01229},
  year={2026}
}

@inproceedings{rt2,
  title={Rt-2: Vision-language-action models transfer web knowledge to robotic control},
  author={Zitkovich, Brianna and Yu, Tianhe and Xu, Sichun and Xu, Peng and Xiao, Ted and Xia, Fei and Wu, Jialin and Wohlhart, Paul and Welker, Stefan and Wahid, Ayzaan and others},
  booktitle={Conference on Robot Learning},
  pages={2165--2183},
  year={2023},
  organization={PMLR}
}

@inproceedings{kim2024openvla,
  title={OpenVLA: An Open-Source Vision-Language-Action Model},
  author={Kim, Moo Jin and Pertsch, Karl and Karamcheti, Siddharth and Xiao, Ted and Balakrishna, Ashwin and Nair, Suraj and Rafailov, Rafael and Foster, Ethan P and Sanketi, Pannag R and Vuong, Quan and others},
  booktitle={8th Annual Conference on Robot Learning},
  year={2024}
}

@inproceedings{octo_2023,
  title={Octo: An Open-Source Generalist Robot Policy},
  author={{Octo Model Team} and Ghosh, Dibya and Walke, Homer and Pertsch, Karl and Black, Kevin and Mees, Oier and Dasari, Sudeep and Hejna, Joey and Xu, Charles and Luo, Jianlan and others},
  booktitle={Proceedings of Robotics: Science and Systems},
  address={Delft, Netherlands},
  year={2024}
}

@article{gr1,
  title={Unleashing Large-Scale Video Generative Pre-training for Visual Robot Manipulation},
  author={Wu, Hongtao and Jing, Ya and Cheang, Chilam and Chen, Guangzeng and Xu, Jiafeng and Li, Xinghang and Liu, Minghuan and Li, Hang and Kong, Tao},
  journal={ICLR},
  year={2024}
}

@inproceedings{rdt,
  title={RDT-1B: a Diffusion Foundation Model for Bimanual Manipulation},
  author={Liu, Songming and Wu, Lingxuan and Li, Bangguo and Tan, Hengkai and Chen, Huayu and Wang, Zhengyi and Xu, Ke and Su, Hang and Zhu, Jun},
  booktitle={The Thirteenth International Conference on Learning Representations},
  year={2024}
}

@article{pi0,
  title={{$\pi_0$}: A Vision-Language-Action Flow Model for General Robot Control},
  author={Black, Kevin and Brown, Noah and Driess, Danny and Esmail, Adnan and Equi, Michael and Finn, Chelsea and Fusai, Niccolo and Groom, Lachy and Hausman, Karol and Ichter, Brian and others},
  journal={arXiv preprint arXiv:2410.24164},
  year={2024}
}

@article{pi05,
  title={{$\pi_{0.5}$}: a Vision-Language-Action Model with Open-World Generalization},
  author={Black, Kevin and Brown, Noah and Darpinian, James and Dhabalia, Karan and Driess, Danny and Esmail, Adnan and Equi, Michael Robert and Finn, Chelsea and Fusai, Niccolo and Galliker, Manuel Y. and others},
  journal={arXiv preprint arXiv:2504.16054},
  year={2025}
}

@article{x-vla,
  title={X-VLA: Soft-prompted Transformer as Scalable Cross-Embodiment Vision-Language-Action Model},
  author={Zheng, Jiandong and Li, Junfeng and Wang, Ziyang and Liu, Dawei and Kang, Xiang and Feng, Yejin and others},
  journal={arXiv preprint arXiv:2510.10274},
  year={2025}
}

@article{llava-vla,
  title={Rethinking the Practicality of Vision-language-action Model: A Comprehensive Benchmark and An Improved Baseline},
  author={Song, Wenxuan and Chen, Jiayi and Sun, Xiaoquan and Lei, Huashuo and Qin, Yikai and Zhao, Wei and Ding, Pengxiang and Zhao, Han and Wang, Tongxin and Hou, Pengxu and others},
  journal={arXiv preprint arXiv:2602.22663},
  year={2026}
}

@inproceedings{du2023universal,
  title={Learning Universal Policies via Text-Guided Video Generation},
  author={Du, Yilun and Yang, Sherry and Dai, Bo and Dai, Hanjun and Nachum, Ofir and Tenenbaum, Josh and Schuurmans, Dale and Abbeel, Pieter},
  booktitle={Advances in Neural Information Processing Systems},
  volume={36},
  pages={9156--9172},
  year={2023}
}

@article{feng2025vidar,
  title={Vidar: Embodied Video Diffusion Model for Generalist Manipulation},
  author={Feng, Yao and Tan, Hengkai and Mao, Xinyi and Xiang, Chendong and Liu, Guodong and Huang, Shuhe and Su, Hang and Zhu, Jun},
  journal={arXiv preprint arXiv:2507.12898},
  year={2025}
}

@inproceedings{zhou2024robodreamer,
  title={RoboDreamer: Learning Compositional World Models for Robot Imagination},
  author={Zhou, Siyuan and Du, Yilun and Chen, Jiaben and Li, Yandong and Yeung, Dit-Yan and Gan, Chuang},
  booktitle={International Conference on Machine Learning},
  pages={61885--61896},
  year={2024}
}

@article{lv2025f1,
  title={F1: A Vision-Language-Action Model Bridging Understanding and Generation to Actions},
  author={Lv, Qi and Kong, Weijie and Li, Hao and Zeng, Jia and Qiu, Zherui and Qu, Delin and Song, Haoming and Chen, Qizhi and Deng, Xiang and Pang, Jiangmiao},
  journal={arXiv preprint arXiv:2509.06951},
  year={2025}
}

@article{zhu2025uwm,
  title={Unified World Models: Coupling Video and Action Diffusion for Pretraining on Large Robotic Datasets},
  author={Zhu, Chuning and Yu, Raymond and Feng, Siyuan and Burchfiel, Benjamin and Shah, Paarth and Gupta, Abhishek},
  journal={arXiv preprint arXiv:2504.02792},
  year={2025}
}

@article{motus,
  title={Motus: A Unified Latent Action World Model},
  author={Bi, Hongzhe and Tan, Hengkai and Xie, Shenghao and Wang, Zeyuan and Huang, Shuhe and Liu, Haitian and Zhao, Ruowen and Feng, Yao and Xiang, Chendong and Rong, Yinze and Zhao, Hongyan and Liu, Hanyu and Su, Zhizhong and Ma, Lei and Su, Hang and Zhu, Jun},
  journal={arXiv preprint arXiv:2512.13030},
  year={2025}
}

@article{dp,
  title={Diffusion policy: Visuomotor policy learning via action diffusion},
  author={Chi, Cheng and Xu, Zhenjia and Feng, Siyuan and Cousineau, Eric and Du, Yilun and Burchfiel, Benjamin and Tedrake, Russ and Song, Shuran},
  journal={The International Journal of Robotics Research},
  pages={02783649241273668},
  year={2023},
  publisher={SAGE Publications Sage UK: London, England}
}

@article{act,
  title        = {Learning Fine-Grained Bimanual Manipulation with Low-Cost Hardware},
  author       = {Zhao, Tony and others},
  journal      = {arXiv preprint arXiv:2304.13705},
  year         = {2023}
}

@article{libero-plus,
    title={LIBERO-Plus: In-depth Robustness Analysis of Vision-Language-Action Models},
    author={Senyu Fei and Siyin Wang and Junhao Shi and Zihao Dai and Jikun Cai and Pengfang Qian and Li Ji and Xinzhe He and Shiduo Zhang and Zhaoye Fei and Jinlan Fu and Jingjing Gong and Xipeng Qiu},
    journal = {arXiv preprint arXiv:2510.13626},
    year={2025},
}

@article{atomvla,
  title={AtomVLA: Scalable Post-Training for Robotic Manipulation via Predictive Latent World Models},
  author={Sun, Xiaoquan and Xu, Zetian and Cao, Chen and Liu, Zonghe and Sun, Yihan and Pang, Jingrui and Zhang, Ruijian and Yang, Zhen and Pang, Kang and He, Dingxin and others},
  journal={arXiv preprint arXiv:2603.08519},
  year={2026}
}

@article{worldvla,
  title={WorldVLA: Towards Autoregressive Action World Model},
  author={Cen, Jun and Yu, Chaohui and Yuan, Hangjie and Jiang, Yuming and Huang, Siteng and Guo, Jiayan and Li, Xin and Song, Yibing and Luo, Hao and Wang, Fan and others},
  journal={arXiv preprint arXiv:2506.21539},
  year={2025}
}

@article{lingbot-va,
  title={Causal World Modeling for Robot Control},
  author={Li, Lin and Zhang, Qihang and Luo, Yiming and Yang, Shuai and Wang, Ruilin and Han, Fei and Yu, Mingrui and Gao, Zelin and Xue, Nan and Zhu, Xing and Shen, Yujun and Xu, Yinghao},
  journal={arXiv preprint arXiv:2601.21998},
  year={2026}
}

@article{fastwam,
  title={Fast-WAM: Do World Action Models Need Test-time Future Imagination?},
  author={Tianyuan Yuan and Zibin Dong and Yicheng Liu and Hang Zhao},
  journal={arXiv preprint arXiv:2603.16666},
  year={2026}
}

@InProceedings{DPFlow,
  author    = {Morimitsu, Henrique and Zhu, Xiaobin and Cesar-Jr., Roberto M. and Ji, Xiangyang and Yin, Xu-Cheng},
  booktitle = {The IEEE/CVF Conference on Computer Vision and Pattern Recognition (CVPR)},
  title     = {{DPFlow}: Adaptive Optical Flow Estimation with a Dual-Pyramid Framework},
  year      = {2025},
}

@article{cronusvla,
  title={CronusVLA: Towards Efficient and Robust Manipulation via Multi-Frame Vision-Language-Action Modeling},
  author={Li, Hao and Yang, Shuai and Chen, Yilun and Tian, Yang and Yang, Xiaoda and Chen, Xinyi and Wang, Hanqing and Wang, Tai and Zhao, Feng and Lin, Dahua and others},
  journal={arXiv preprint arXiv:2506.19816},
  year={2025}
}

@article{memoryvla,
  title={MemoryVLA: Perceptual-Cognitive Memory in Vision-Language-Action Models for Robotic Manipulation},
  author={Shi, H. and Xie, B. and Liu, Y. and Sun, L. and Liu, F. and Wang, T. and Zhou, E. and Fan, H. and Zhang, X. and Huang, G.},
  journal={arXiv preprint arXiv:2508.19236},
  year={2025}
}

@article{memer,
  title={MemER: Scaling Up Memory for Robot Control via Experience Retrieval},
  author={Sridhar, A. and Pan, J. and Sharma, S. and Finn, C.},
  journal={arXiv preprint arXiv:2510.20328},
  year={2025}
}

@article{sam2act,
  title={{SAM2Act}: Integrating Visual Foundation Model with a Memory Architecture for Robotic Manipulation},
  author={Fang, H. and Grotz, M. and Pumacay, W. and Wang, Y. R. and Fox, D. and Krishna, R. and Duan, J.},
  journal={arXiv preprint arXiv:2501.18564},
  year={2025}
}

@article{cogact,
  title={{CogACT}: A Foundational Vision-Language-Action Model for Synergizing Cognition and Action in Robotic Manipulation},
  author={Li, Q. and Liang, Y. and Wang, Z. and Luo, L. and Chen, X. and Liao, M. and Wei, F. and Deng, Y. and Xu, S. and Zhang, Y. and others},
  journal={arXiv preprint arXiv:2411.19650},
  year={2024}
}

@article{wang2026trust,
  title={When to Trust Imagination: Adaptive Action Execution for World Action Models},
  author={Wang, Rui and Zhang, Yue and Lin, Jiehong and Luo, Kuncheng and Wang, Jianan and Wang, Zhongrui and Qi, Xiaojuan},
  journal={arXiv preprint arXiv:2605.06222},
  year={2026}
}

@article{team2026motubrain,
  title={MotuBrain: An Advanced World Action Model for Robot Control},
  author={Team, MotuBrain and Xiang, Chendong and Bao, Fan and Liu, Haitian and Tan, Hengkai and Bi, Hongzhe and Li, James and Liu, Jiabao and Pang, Jingrui and Jing, Kiro and others},
  journal={arXiv preprint arXiv:2604.27792},
  year={2026}
}

\clearpage
\appendix

\begin{center}
{\large\bfseries Supplementary Material}
\end{center}
\phantomsection
\label{app:overview}

This supplementary material provides additional details on the implementation and evaluation of \method. Appendix~\ref{app:method_details} describes the main components of our framework, including latent action learning, skill modeling, memory construction, training stages, and inference. Appendix~\ref{app:real_world_tasks} presents the real world evaluation protocol, perturbation settings, and task definitions. Appendix~\ref{app:baselines} summarizes the baseline models used for comparison. These details complement the main paper and provide a clearer view of our experimental setup and implementation choices.

\section{Method Details}
\label{app:method_details}This section provides the implementation details omitted from the main Method section. We use $z^l$ for low-level latent actions and $z^h$ for high-level latent actions or skill latents throughout the appendix. The details below preserve the same three-stage training pipeline as the main paper: offline low-level latent-action tokenizer learning, hierarchical skill discovery and latent-policy pretraining, and action grounding with gated memory.
\subsection{Separation Between Offline Supervision and Online Inputs}
\label{app:offline_online}

\method\ uses rich motion supervision during training but keeps inference causal. Table~\ref{tab:offline_online_signals} summarizes which signals are used in each stage.

\begin{table}[h]
\centering
\small
\setlength{\tabcolsep}{5pt}
\renewcommand{\arraystretch}{1.10}
\caption{Signals used during training and inference. Optical flow and future observations are used only to construct supervision for latent-action learning; they are not required by the deployed policy.}
\label{tab:offline_online_signals}
\begin{tabular}{lcccc}
\toprule
\textbf{Signal} & \textbf{Stage I} & \textbf{Stage II} & \textbf{Stage III} & \textbf{Inference} \\
\midrule
RGB observations $o_t$ & \checkmark & \checkmark & \checkmark & \checkmark \\
Proprioception $p_t$ & \checkmark & \checkmark & \checkmark & \checkmark \\
Instruction $\ell$ & \checkmark & \checkmark & \checkmark & \checkmark \\
Optical flow $\mathbf{\Phi}_t$ & \checkmark & supervision only & -- & -- \\
Action annotations $a_t$ & optional & -- & \checkmark & -- \\
Low-level latents $Z^l$ & output & supervision & supervision & predicted \\
High-level skills $\bar z_t^h$ & -- & supervision & auxiliary supervision & predicted \\
Boundary labels $\bar b_t$ & -- & supervision & gate supervision & predicted \\
External memory $\mathcal{M}_t$ & -- & disabled & \checkmark & \checkmark \\
\bottomrule
\end{tabular}
\end{table}

Stage I uses optical flow to learn a compact low-level latent-action space. Stage II uses the extracted $Z^l$ sequences to discover high-level skills and pretrain the planner and executor without memory. Stage III activates the gated external memory module and grounds the latent policy into executable actions. During inference, the policy receives only RGB observations, proprioception, instruction, and the current memory bank.

\subsection{Low-Level Latent-Action Tokenizer Details}
\label{app:tokenizer_details}

\paragraph{Optical-flow preprocessing.}
For each view $v$, DPFlow produces a dense flow field
\[
\Phi_t^{(v)}=\mathrm{DPFlow}(I_t^{(v)},I_{t+1}^{(v)}).
\]
Before the flow is passed to the encoder, its horizontal and vertical components are normalized by the image width and height:
\begin{equation}
\widetilde{\Phi}_t^{(v)}(x,y)
=
\left[
\frac{\Phi_{t,x}^{(v)}(x,y)}{W_v},
\frac{\Phi_{t,y}^{(v)}(x,y)}{H_v}
\right],
\label{eq:app_flow_norm}
\end{equation}
where $W_v$ and $H_v$ denote the resolution of view $v$. This normalization keeps flow magnitudes comparable across views.

\paragraph{Multi-view fusion.}
Each view is encoded independently and receives a view embedding $e_v$:
\begin{equation}
m_t^{(v)}=E_{\mathrm{flow}}(\widetilde{\Phi}_t^{(v)})+e_v,
\qquad
s_t^{(v)}=E_{\mathrm{vis}}(I_t^{(v)})+e_v.
\label{eq:app_view_encoding}
\end{equation}
The motion and semantic features are fused across views:
\begin{equation}
m_t=\mathrm{Fuse}_{\mathrm{mot}}(\{m_t^{(v)}\}_{v=1}^{V}),
\qquad
s_t=\mathrm{Fuse}_{\mathrm{sem}}(\{s_t^{(v)}\}_{v=1}^{V}).
\label{eq:app_view_fusion}
\end{equation}
Together with proprioception and instruction features, these features form the context $c_t$ used by the low-level posterior in Eq.~\eqref{eq:low_latent_posterior}.

\paragraph{Action-annotation masking.}
The tokenizer can use both action-labeled robot trajectories and action-free videos. Let $\mathbb{I}^{\mathrm{act}}_t\in\{0,1\}$ indicate whether $a_t$ is available. The action-alignment loss is applied only to labeled transitions:
\begin{equation}
\mathcal{L}_{\mathrm{align}}
=
\frac{1}{\sum_{t=1}^{T-1}\mathbb{I}^{\mathrm{act}}_t}
\sum_{t=1}^{T-1}
\mathbb{I}^{\mathrm{act}}_t
\left\|
D_{\mathrm{align}}(z_t^l,o_t,p_t)-a_t
\right\|_2^2.
\label{eq:app_action_mask}
\end{equation}
Thus, action-free videos contribute through flow reconstruction and KL regularization, while action-labeled robot data additionally aligns $z^l$ with executable controls.

\paragraph{Offline latent extraction.}
After Stage I, the tokenizer is frozen and applied offline to all training trajectories and videos. For each sequence, we store
\[
Z^l=(z_1^l,\ldots,z_{T-1}^l).
\]
These stored latents are used as pseudo-labels in Stage II and Stage III. DPFlow is not called during policy training after this extraction step, and it is not used during inference.

\subsection{High-Level Latent-Action Learning}
\label{app:skill_learning_details}

The main text summarizes high-level latent-action learning with the operator $\mathrm{Chunk}_s$. Here we provide the explicit boundary scoring, pooling, and loss terms.

\paragraph{Boundary scoring.}
Let $Z^{(0)}=Z^l$ and let $L_s$ be the length of $Z^{(s)}$. At hierarchy stage $s$, each token is encoded as $h_i^{(s)}=E_s(z_i^{(s)})$. We compute normalized query and key features,
\[
\hat q_i^{(s)}=\frac{W_q^{(s)}h_i^{(s)}}{\|W_q^{(s)}h_i^{(s)}\|_2},
\qquad
\hat k_i^{(s)}=\frac{W_k^{(s)}h_i^{(s)}}{\|W_k^{(s)}h_i^{(s)}\|_2},
\]
and define a dissimilarity score
\begin{equation}
r_i^{(s)}
=
\begin{cases}
1, & i=1,\\
\frac{1}{2}\left(1-(\hat q_{i-1}^{(s)})^\top \hat k_i^{(s)}\right), & i>1.
\end{cases}
\label{eq:app_boundary_score}
\end{equation}
The boundary indicator is
\begin{equation}
b_i^{(s)}
=
\begin{cases}
1, & i=1,\\
\mathbb{I}\!\left[r_i^{(s)}\ge \delta_s\right], & i>1,
\end{cases}
\label{eq:app_skill_boundary}
\end{equation}
where $b_i^{(s)}=1$ means that token $i$ starts a new segment.

\paragraph{Variable-length segment pooling.}
Let the ordered boundary indices at stage $s$ be
\[
\mathcal{B}^{(s)}=\{i_j^{(s)}\}_{j=1}^{L_{s+1}}=\{i\mid b_i^{(s)}=1\},
\]
with $i_{L_{s+1}+1}^{(s)}=L_s+1$. The $j$-th segment is
\[
\mathcal{I}_j^{(s)}=\{i\mid i_j^{(s)}\le i<i_{j+1}^{(s)}\}.
\]
We summarize each segment by attention pooling:
\begin{equation}
\alpha_{j,i}^{(s)}
=
\frac{\exp((w_s)^\top h_i^{(s)})}
{\sum_{k\in\mathcal{I}_j^{(s)}}\exp((w_s)^\top h_k^{(s)})},
\qquad
z_j^{(s+1)}
=
\sum_{i\in\mathcal{I}_j^{(s)}}\alpha_{j,i}^{(s)}h_i^{(s)}.
\label{eq:app_skill_chunk}
\end{equation}
After $H$ hierarchy stages, $Z^h=Z^{(H)}=(z_1^h,\ldots,z_S^h)$, where $S=L_H\ll T$.

\paragraph{Skill-discovery loss.}
The hierarchy is trained with
\begin{equation}
\mathcal{L}_{\mathrm{skill}}
=
\mathcal{L}_{\mathrm{next}}
+
\lambda_m\mathcal{L}_{\mathrm{motion}}
+
\lambda_r\mathcal{L}_{\mathrm{ratio}}
+
\lambda_c\mathcal{L}_{\mathrm{cons}}.
\label{eq:app_skill_loss_full}
\end{equation}
The next-latent loss predicts the next low-level latent action:
\begin{equation}
\hat z_{t+1}^l=P_\omega(z_t^l,\bar z_t^h),
\qquad
\mathcal{L}_{\mathrm{next}}
=
\frac{1}{T-2}
\sum_{t=1}^{T-2}
\left\|\hat z_{t+1}^l-z_{t+1}^l\right\|_1.
\label{eq:app_next_loss}
\end{equation}
To preserve motion semantics, the frozen flow decoder reconstructs flow from the predicted latent:
\begin{equation}
\hat{\mathbf{\Phi}}_{t+1}^{\mathrm{pred}}
=
D_{\mathrm{flow}}(\hat z_{t+1}^l,o_{t+1}),
\qquad
\mathcal{L}_{\mathrm{motion}}
=
\frac{1}{V(T-2)}
\sum_{t=1}^{T-2}\sum_{v=1}^{V}
\left\|
\hat{\Phi}_{t+1}^{(v),\mathrm{pred}}-
\Phi_{t+1}^{(v)}
\right\|_1.
\label{eq:app_motion_loss}
\end{equation}
The ratio loss prevents degenerate boundary patterns:
\begin{equation}
\mathcal{L}_{\mathrm{ratio}}
=
\sum_{s=0}^{H-1}
\left(
\frac{1}{L_s}
\sum_{i=1}^{L_s}b_i^{(s)}
-
\rho_s
\right)^2,
\label{eq:app_ratio_loss}
\end{equation}
where $\rho_s$ is the target boundary ratio. The consistency term encourages tokens in the same segment to share a coherent representation:
\begin{equation}
\mathcal{L}_{\mathrm{cons}}
=
\sum_{s=0}^{H-1}
\frac{1}{L_s}
\sum_{j=1}^{L_{s+1}}
\sum_{i\in\mathcal{I}_j^{(s)}}
\left\|
R_sh_i^{(s)}-z_j^{(s+1)}
\right\|_2^2.
\label{eq:app_cons_loss}
\end{equation}

\subsection{Skill Boundary Unfolding and Pseudo-Label Construction}
\label{app:skill_unfolding}

Since higher hierarchy stages operate on progressively shorter sequences, we maintain an index map $\psi_s$ from each stage-$s$ token to its starting timestep in the original low-level sequence. At the bottom level,
\begin{equation}
\psi_0(i)=i,
\qquad i=1,\ldots,L_0.
\label{eq:app_index_map_init}
\end{equation}
If the boundary indices at stage $s$ are $\mathcal{B}^{(s)}=\{i_j^{(s)}\}_{j=1}^{L_{s+1}}$, the next-stage map is
\begin{equation}
\psi_{s+1}(j)=\psi_s(i_j^{(s)}),
\qquad j=1,\ldots,L_{s+1}.
\label{eq:app_index_map_update}
\end{equation}
After $H$ stages, the final boundary indicator at the original temporal resolution is
\begin{equation}
\bar b_t
=
\mathbb{I}\!\left[t\in\{\psi_H(j)\}_{j=1}^{S}\right],
\qquad
 t=1,\ldots,T-1,
\label{eq:app_final_boundary}
\end{equation}
with $\bar b_1=1$. The segment index and per-timestep high-level target are
\begin{equation}
\kappa_t=\sum_{\tau=1}^{t}\bar b_\tau,
\qquad
\bar z_t^h=z_{\kappa_t}^h,
\qquad
 t=1,\ldots,T-1.
\label{eq:app_skill_target}
\end{equation}
For action-chunk policy learning, the low-level target at timestep $t$ is
\begin{equation}
\mathbf{Z}_{t:t+K-1}^{l}
=
(z_t^l,\ldots,z_{t+K-1}^l).
\label{eq:app_low_chunk_target}
\end{equation}
When $t+K-1>T-1$, we use the valid suffix and mask invalid positions in the executor loss.

\subsection{Qwen3-VL-4B-Instruct Planner and Memory Interface}
\label{app:qwen_planner}

The planner is instantiated with Qwen3-VL-4B-Instruct and is responsible for high-level skill prediction and memory-conditioned boundary estimation. It is not itself the memory bank.

\paragraph{Planner input formatting.}
At timestep $t$, the planner receives the current multi-view RGB observation $o_t$, instruction $\ell$, a projected proprioceptive summary $P_p(p_t)$, and a projected memory context $P_m(c_t^m)$. The planner hidden state is
\begin{equation}
h_t^{\mathrm{plan}}
=
\mathrm{QwenPlan}_{\theta}
(o_t,\ell,P_p(p_t),P_m(c_t^m)).
\label{eq:app_qwen_plan}
\end{equation}
Continuous heads predict the high-level latent and boundary score:
\begin{equation}
\hat z_t^h=H_z(h_t^{\mathrm{plan}}),
\qquad
\hat b_t=\sigma(H_b(h_t^{\mathrm{plan}})).
\label{eq:app_qwen_heads}
\end{equation}

\paragraph{Memory retrieval.}
The external memory bank is a set of continuous skill-level tokens, $\mathcal{M}_t=\{m_i\}_{i=1}^{N_t}$. Given $x_t=E_\theta(o_t,p_t,\ell)$, the retrieved context is
\begin{equation}
c_t^m
=
\mathrm{Attn}(W_qx_t,W_k\mathcal{M}_t,W_v\mathcal{M}_t),
\label{eq:app_mem_retrieval}
\end{equation}
with $c_t^m=0$ if $\mathcal{M}_t$ is empty. The read gate forms the memory-adapted state:
\begin{equation}
\alpha_t^r
=
\sigma(G_r(x_t,c_t^m)),
\qquad
\tilde{x}_t
=
x_t+\alpha_t^r W_m c_t^m.
\label{eq:app_mem_read}
\end{equation}

\paragraph{Memory writing.}
After the planner predicts $\hat z_t^h$ and $\hat b_t$, the executor produces
\[
\hat{\mathbf{Z}}_{t:t+K-1}^{l}
=
\pi_\theta^{\mathrm{exec}}(\tilde{x}_t,\hat z_t^h).
\]
The write gate and memory token are
\begin{equation}
\alpha_t^w
=
\sigma(G_w(\tilde{x}_t,\hat z_t^h,\hat b_t)),
\qquad
\gamma_t
=
\Gamma_\psi\!\left(
\tilde{x}_t,
\hat z_t^h,
\mathrm{Pool}(\hat{\mathbf{Z}}_{t:t+K-1}^{l})
\right).
\label{eq:app_mem_token}
\end{equation}
The memory bank is updated by
\begin{equation}
\mathcal{M}_{t+1}
=
\begin{cases}
U_\psi(\mathcal{M}_t,\gamma_t), & \alpha_t^w>\eta,\\
\mathcal{M}_t, & \mathrm{otherwise}.
\end{cases}
\label{eq:app_mem_update}
\end{equation}
The update operator $U_\psi$ appends $\gamma_t$ to the bank and compresses the bank if it exceeds a fixed budget $N_{\max}$.

\subsection{Training Details for the Three Stages}
\label{app:training_details}

\paragraph{Stage I.}
Stage I trains only the low-level tokenizer. The output is a frozen tokenizer and a set of offline $Z^l$ pseudo-labels. The planner, executor, action decoder, and memory gates are not trained in this stage.

\paragraph{Stage II.}
Stage II learns high-level pseudo-labels and pretrains the planner and executor without external memory. We set $\mathcal{M}_t=\emptyset$, $c_t^m=0$, and $\tilde{x}_t=x_t$. The loss is
\begin{equation}
\mathcal{L}_{\mathrm{latent}}
=
\lambda_h\mathcal{L}_{\mathrm{plan}}
+
\lambda_l\mathcal{L}_{\mathrm{exec}}
+
\lambda_b\mathcal{L}_{\mathrm{bd}},
\label{eq:app_latent_loss}
\end{equation}
where
\begin{equation}
\mathcal{L}_{\mathrm{plan}}
=
\frac{1}{T-1}
\sum_{t=1}^{T-1}
\left\|
\hat z_t^h-\bar z_t^h
\right\|_2^2,
\label{eq:app_plan_loss}
\end{equation}
\begin{equation}
\mathcal{L}_{\mathrm{exec}}
=
\frac{1}{K(T-K)}
\sum_{t=1}^{T-K}
\left\|
\hat{\mathbf{Z}}_{t:t+K-1}^{l}
-
\mathbf{Z}_{t:t+K-1}^{l}
\right\|_2^2,
\label{eq:app_exec_loss}
\end{equation}
and
\begin{equation}
\mathcal{L}_{\mathrm{bd}}
=
\frac{1}{T-1}
\sum_{t=1}^{T-1}
\mathrm{BCE}(\hat b_t,\bar b_t).
\label{eq:app_bd_loss}
\end{equation}

\paragraph{Stage III.}
Stage III activates memory and fine-tunes the policy on action-labeled robot demonstrations. The full objective is
\begin{equation}
\mathcal{L}_{\mathrm{ft}}
=
\mathcal{L}_{\mathrm{act}}
+
\alpha_h\mathcal{L}_{\mathrm{plan}}
+
\alpha_l\mathcal{L}_{\mathrm{exec}}
+
\alpha_b\mathcal{L}_{\mathrm{bd}}
+
\alpha_m\mathcal{L}_{\mathrm{gate}}.
\label{eq:app_ft_loss}
\end{equation}
For deterministic action prediction,
\begin{equation}
\mathcal{L}_{\mathrm{act}}
=
\left\|
\hat{\mathbf{a}}_{t:t+K-1}
-
\mathbf{a}_{t:t+K-1}
\right\|_2^2.
\label{eq:app_action_loss_l2}
\end{equation}
For stochastic policies, it is replaced by
\begin{equation}
\mathcal{L}_{\mathrm{act}}
=
-
\log
\pi_\theta(\mathbf{a}_{t:t+K-1}\mid o_t,p_t,\ell,\mathcal{M}_t).
\label{eq:app_action_loss_nll}
\end{equation}
The gate loss is
\begin{equation}
\mathcal{L}_{\mathrm{gate}}
=
\frac{1}{T-1}
\sum_{t=1}^{T-1}
\left[
\mathrm{BCE}(\alpha_t^w,\bar b_t)
+
\lambda_r\|\alpha_t^r\|_1
+
\lambda_w\|\alpha_t^w\|_1
\right].
\label{eq:app_gate_loss}
\end{equation}
The write gate is supervised by the discovered boundary label $\bar b_t$, while the read gate is learned through downstream action and latent prediction losses with sparsity regularization.

\paragraph{Teacher-forced memory warmup.}
To stabilize Stage III, memory updates can be initialized with the discovered boundary label rather than the predicted write gate. During warmup,
\begin{equation}
\mathcal{M}_{t+1}^{\mathrm{teach}}
=
\begin{cases}
U_\psi(\mathcal{M}_{t}^{\mathrm{teach}},\gamma_t^{\mathrm{teach}}), & \bar b_t=1,\\
\mathcal{M}_{t}^{\mathrm{teach}}, & \mathrm{otherwise},
\end{cases}
\label{eq:app_teacher_memory}
\end{equation}
where
\begin{equation}
\gamma_t^{\mathrm{teach}}
=
\Gamma_\psi\!\left(
 x_t,
 \bar z_t^h,
 \mathrm{Pool}(\mathbf{Z}_{t:t+K-1}^{l})
\right).
\label{eq:app_teacher_token}
\end{equation}
After warmup, memory writing is controlled by $\alpha_t^w$ as in Eq.~\eqref{eq:app_mem_update}.

\subsection{Inference Pipeline.}
\label{app:inference}

At inference time, \method\ does not generate future videos and does not estimate optical flow. At each decision step, it executes the following causal procedure:
\begin{enumerate}[leftmargin=*]
    \item Receive current RGB observations $o_t$, proprioception $p_t$, instruction $\ell$, and memory bank $\mathcal{M}_t$.
    \item Compute $x_t=E_\theta(o_t,p_t,\ell)$ and retrieve $c_t^m$ from $\mathcal{M}_t$.
    \item Form $\tilde{x}_t=x_t+\alpha_t^rW_mc_t^m$.
    \item Use the Qwen3-VL-4B-Instruct planner to predict $\hat z_t^h$ and $\hat b_t$.
    \item Use the executor to generate $\hat{\mathbf{Z}}_{t:t+K-1}^{l}$.
    \item Decode $\hat{\mathbf{a}}_{t:t+K-1}=D_{\mathrm{act}}(\hat{\mathbf{Z}}_{t:t+K-1}^{l},\tilde{x}_t)$.
    \item If $\alpha_t^w>\eta$, write $\gamma_t$ into the memory bank.
\end{enumerate}
This procedure uses only current observations and stored memory, preserving the standard causal interface of action-chunking robot policies.

\clearpage
\section{Real-World Setting Details}
\label{app:real_world_tasks}

\subsection{Generalization Setting}
We provide the \method\ definition of the \textbf{GE} setting used in our real-world evaluation. Depending on the task, one or more of the following perturbations may be introduced:
\begin{itemize}[leftmargin=*]
    \item \textbf{Object position variation:} The initial positions of target objects and receptacles are changed within the workspace.
    \item \textbf{Unseen distractor objects:} Novel objects that are not present in the training demonstrations are placed near the target objects, resulting in a more cluttered scene.
    \item \textbf{Target layout and height variation:} The spatial layout of task-relevant objects is changed, and target objects or receptacles may be placed at different heights using support structures.
    \item \textbf{Lighting variation:} The illumination conditions are changed to evaluate robustness under different lighting environments.
    \item \textbf{Instruction variation:} The original task instructions are replaced with semantically equivalent paraphrases to evaluate robustness to language variation.
\end{itemize}

\subsection{Task Descriptions}
\noindent\textbf{Easy tasks.} Easy tasks are short-horizon single-arm manipulation tasks involving basic actions such as reaching, grasping, picking, and placing.
\begin{itemize}[leftmargin=*]
    \item \textbf{Stack bowls}: Pick up the yellow bowl and place it on top of the green bowl.
    \item \textbf{Hang cup}: Pick up the cup and hang it on the mug rack.
    \item \textbf{Put fruit into basket}: Pick up the fruit and place it into the basket.
    \item \textbf{Press button}: Press the target button until it is activated.
\end{itemize}
\noindent\textbf{Medium tasks.} Medium tasks require basic bimanual coordination, while the overall task horizon remains moderate and the manipulation procedure is still relatively structured.
\begin{itemize}[leftmargin=*]
    \item \textbf{Stack three bowls}: Pick up the yellow bowl and the pink bowl sequentially, and stack both of them on top of the green bowl.
    \item \textbf{Fold towel}: Grasp one side of the towel and fold it over to the other side along the instructed direction.
    \item \textbf{Place plate}: Pick up the plate and place it onto the plate rack.
    \item \textbf{Press two buttons}: Press the two target buttons in the required order until they are activated.
\end{itemize}

\noindent\textbf{Hard tasks.} Hard tasks involve longer-horizon bimanual manipulation and more complex object interactions, and typically require more precise coordination across multiple steps.
\begin{itemize}[leftmargin=*]
    \item \textbf{Make breakfast}: Pick up one bread slice and place it on the plate, then place a ham slice on top of it, and finally place another bread slice on top to complete the stack.
    \item \textbf{Place two plates}: Pick up two plates sequentially and place both of them onto the plate rack.
\end{itemize}
\label{app:realworld_full_results}

\clearpage

\begin{figure}[h]
    \centering
    \includegraphics[width=1.0\textwidth,height=0.99\textheight,keepaspectratio]{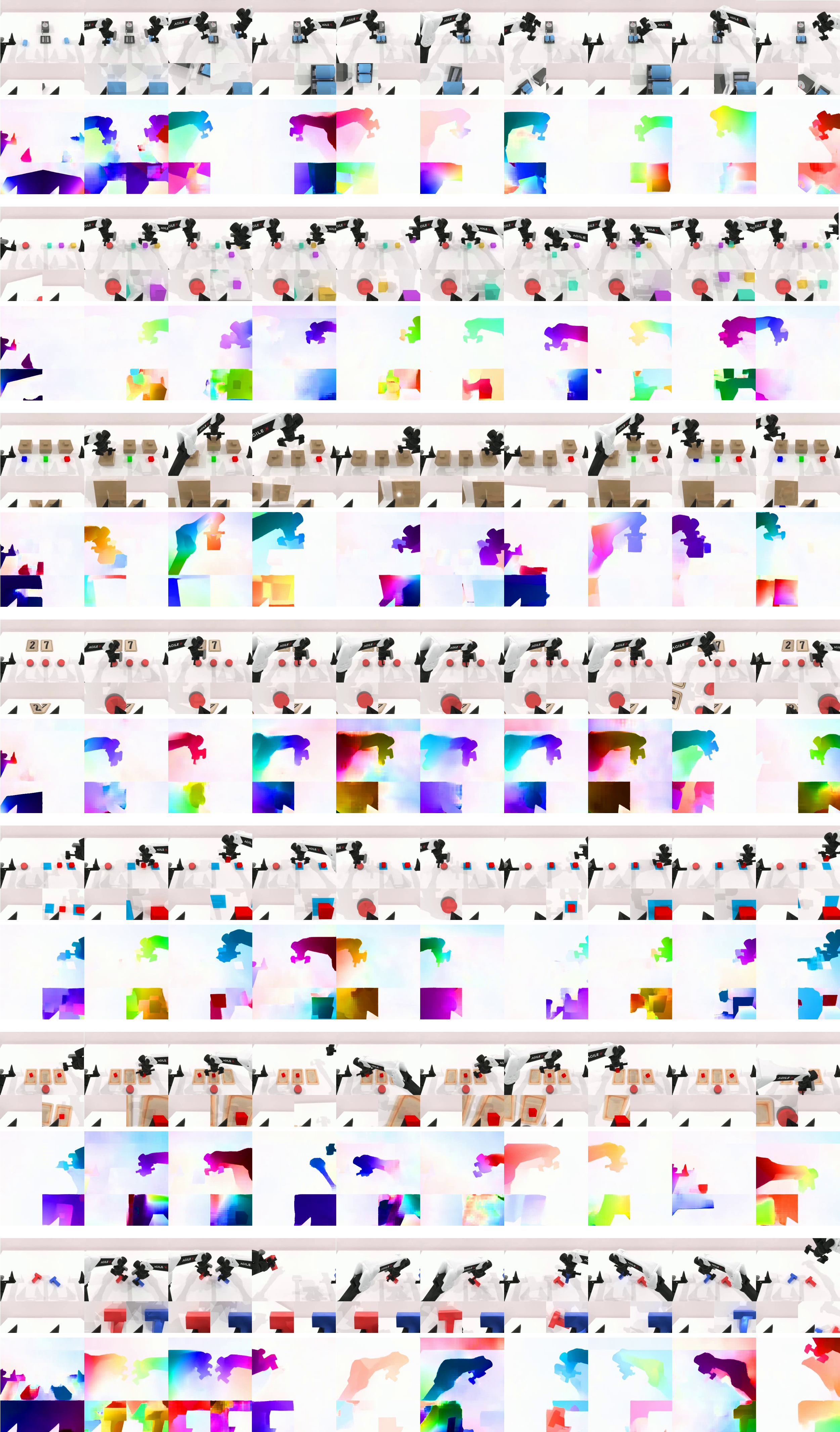}
    \caption{\textbf{RMBench tasks rollout and DPFlow visualization.}}
    \label{fig:appendix_RMBench}
\end{figure}

\clearpage

\begin{figure}[h]
    \centering
    \includegraphics[width=1.0\textwidth]{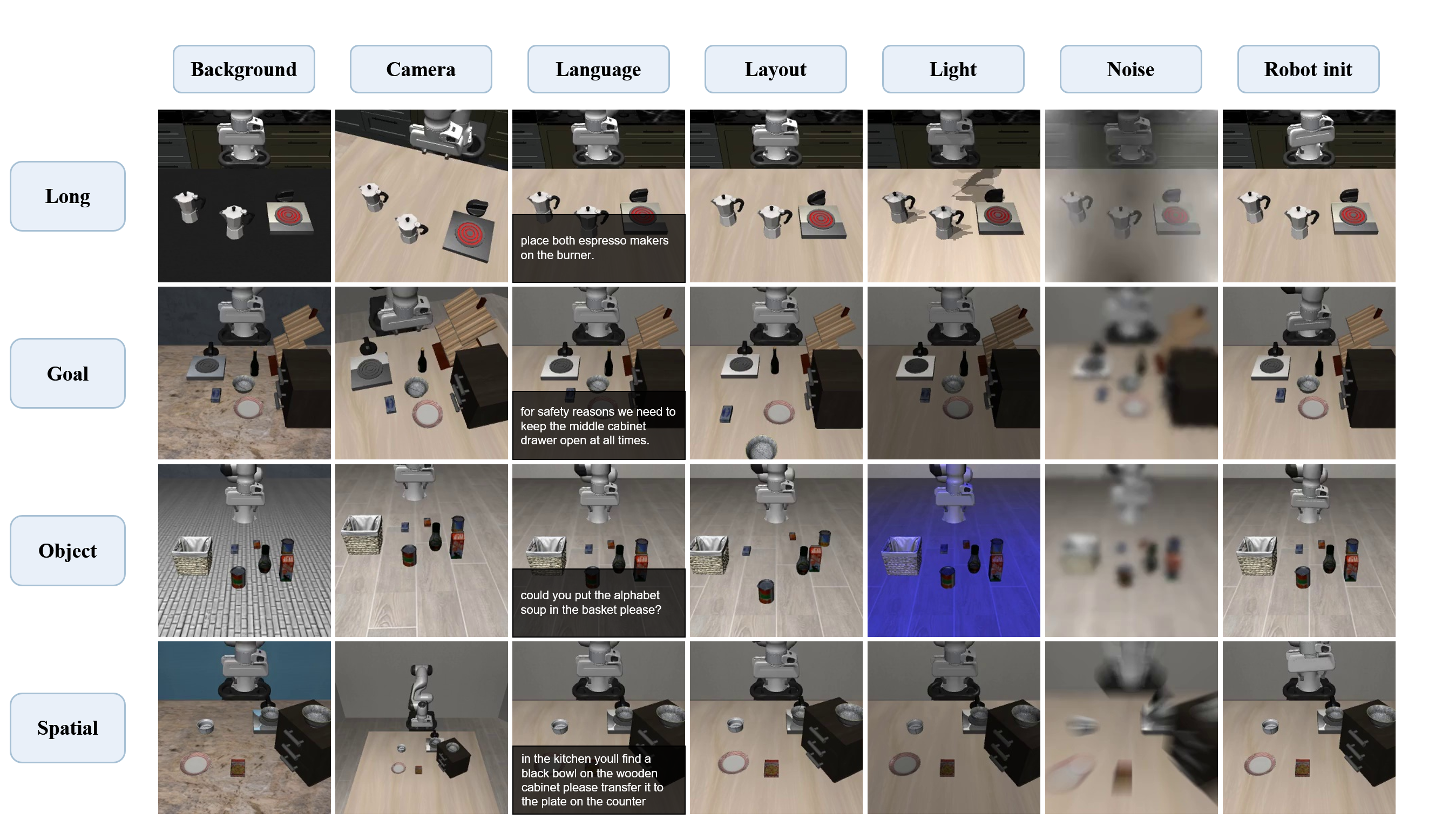}
    \caption{\textbf{LIBERO-Plus tasks rollout visualization.} seven perturbation types in the LIBERO-PLUS benchmark, used to evaluate robustness}
    \label{fig:libero_plus}
\end{figure}

\section{Baselines}
\label{app:baselines}
\noindent\textbf{DP:} Diffusion Policy is a diffusion-based visuomotor imitation learning method that represents robot actions as a conditional denoising process. It predicts action sequences conditioned on visual observations and executes them in a receding-horizon manner, enabling expressive multi-modal action generation for contact-rich manipulation.

\noindent\textbf{ACT:} Action Chunking Transformer is an imitation learning policy originally developed for fine-grained bimanual manipulation. Instead of predicting a single action at each step, ACT generates temporally coherent action chunks with a Transformer-based policy and applies temporal ensembling to reduce compounding errors and improve execution smoothness.

\noindent\textbf{X-VLA:} X-VLA is a soft-prompted flow-matching VLA framework designed for cross-embodiment robot learning. It introduces learnable prompt embeddings to encode robot- and dataset-specific variations, allowing a shared Transformer policy to adapt across heterogeneous sensors, action spaces, and robotic platforms with limited additional parameters.

\noindent\textbf{MEM-0:} MEM-0 is a modular memory-aware manipulation policy introduced for memory-dependent robotic tasks. It explicitly incorporates memory components to retain task-relevant historical observations, making it suitable for evaluating manipulation scenarios where the correct action depends on past states rather than only the current visual input.

\noindent\textbf{OpenVLA:} OpenVLA is a 7B-parameter open-source vision-language-action model trained on 970K real-world robot demonstrations from the Open X-Embodiment dataset. It builds on a pretrained vision-language backbone with DINOv2 and SigLIP visual encoders and a Llama-2 language model, and predicts discretized robot actions from image observations and language instructions.

\noindent\textbf{OpenVLA-OFT:} OpenVLA-OFT is an optimized fine-tuning variant of OpenVLA. It replaces the original token-by-token action generation with a faster and more control-oriented recipe that combines parallel decoding, action chunking, continuous action regression, and an L1 objective, substantially improving inference speed and downstream manipulation success.

\noindent\textbf{SpatialVLA:} SpatialVLA is a spatially enhanced VLA model trained on large-scale real-robot episodes. It augments visual-language-action modeling with Ego3D Position Encoding for explicit 3D spatial grounding and Adaptive Action Grids for discretizing spatial robot motions, improving cross-robot transfer and manipulation in geometry-sensitive tasks.

\noindent\textbf{\boldmath$\pi_0$:} $\pi_0$ is a generalist vision-language-action flow model built on top of a pretrained VLM. It uses flow matching to generate continuous action trajectories and is trained on diverse data from multiple robotic platforms, including single-arm, dual-arm, and mobile manipulation settings, enabling language-conditioned dexterous control and fine-tuning to new skills.

\noindent\textbf{\boldmath$\pi_{0.5}$:} $\pi_{0.5}$ extends $\pi_0$ toward open-world generalization by co-training on heterogeneous robot and non-robot data. In addition to low-level action prediction, it incorporates multimodal supervision such as semantic predictions, object information, and high-level task cues, enabling long-horizon manipulation in unfamiliar real-world environments.

\noindent\textbf{NORA-1.5:} NORA-1.5 builds upon the NORA VLA backbone and augments it with a flow-matching action expert. It further applies reward-guided post-training using world-model-based and action-based preference rewards, improving action reliability and task success across simulation and real-robot settings.

\noindent\textbf{AtomVLA:} AtomVLA is a subtask-aware VLA post-training framework for long-horizon robotic manipulation. It decomposes high-level instructions into atomic subtasks and uses a predictive latent world model to evaluate candidate action chunks, enabling scalable offline policy optimization and reducing compounding errors during multi-step execution.

\noindent\textbf{UniVLA:} UniVLA is a unified vision-language-action model that represents visual observations, language instructions, and robot actions as token sequences within a single autoregressive framework. By jointly modeling perception, world dynamics, and action generation, it aims to improve long-horizon policy learning and multimodal grounding for robotic manipulation.

\noindent\textbf{WorldVLA:} WorldVLA is an autoregressive action world model that integrates VLA policy learning and world modeling into one framework. It jointly reasons about robot actions and future visual states, allowing the model to use predicted environmental dynamics as an intermediate structure for more temporally consistent action generation.

\noindent\textbf{FAST-WAM:} FAST-WAM is a fast world-action model that studies whether explicit future imagination is necessary at test time. It retains video-modeling objectives during training to learn stronger world representations, but skips future-frame generation during inference, achieving competitive manipulation performance with substantially lower control latency.

\noindent\textbf{RIPT-VLA:} RIPT-VLA is an interactive post-training paradigm for pretrained Vision-Language-Action models. It fine-tunes VLA policies through online environment rollouts with sparse binary success rewards, using dynamic rollout sampling and leave-one-out advantage estimation to improve task adaptation and robustness beyond supervised imitation.

\noindent\textbf{HoloBrain-0:} HoloBrain-0 is a comprehensive Vision-Language-Action framework for robotic manipulation that explicitly incorporates robot embodiment priors, such as multi-view camera parameters and kinematic descriptions. It follows a scalable pre-train then post-train paradigm and serves as a strong baseline for robust action generation in simulation and real-world manipulation benchmarks.

\end{document}